# Performance of GPT-5 Frontier Models in Ophthalmology Question Answering

**Fares Antaki**, MDCM[1,2,3,4]; **David Mikhail**[5], BSc; **Daniel Milad**, MD[2,3,6]; **Danny A Mammo**, MD[1]; **Sumit Sharma**, MD[1,7]; **Sunil K Srivastava**, MD[1]; **Bing Yu Chen**, MDCM[8]; **Samir Touma**, MDCM[2,3,6]; **Mertcan Sevgi**, MD[9,10]; **Jonathan El-Khoury**, MD[2,3,6]; **Pearse A Keane**, MD[9,10]; **Qingyu Chen**, PhD[11]; **Yih Chung Tham**, PhD[12,13,14]; **Renaud Duval**, MD[2,6]

1. Cole Eye Institute, Cleveland Clinic, Cleveland, OH, USA
2. Department of Ophthalmology, University of Montreal, Montreal, Quebec, Canada
3. Department of Ophthalmology, Centre Hospitalier de l'Universite de Montreal, Montreal, Quebec, Canada
4. The CHUM School of Artificial Intelligence in Healthcare (SAIH), Centre Hospitalier de l'Université de Montréal (CHUM), Montreal, Quebec, Canada
5. Temerty Faculty of Medicine, University of Toronto, Toronto, Ontario, Canada
6. Department of Ophthalmology, Hopital Maisonneuve-Rosemont, Montreal, Quebec, Canada
7. Cleveland Clinic Lerner College of Medicine of Case Western Reserve University, Cleveland, OH, USA
8. Neurological Institute, Cleveland Clinic, Cleveland, Ohio, USA
9. Institute of Ophthalmology, University College London, London, UK
10. NIHR Biomedical Research Centre at Moorfields, Moorfields Eye Hospital NHS Foundation Trust, London, UK
11. Department of Biomedical Informatics and Data Science, Yale School of Medicine, Yale University, New Haven, Connecticut, USA
12. Centre for Innovation and Precision Eye Health, Department of Ophthalmology, Yong Loo Lin School of Medicine, National University of Singapore, Singapore
13. Singapore Eye Research Institute, Singapore National Eye Centre, Singapore
14. Eye Academic Clinical Program, Duke NUS Medical School, Singapore

**Correspondence:** Fares Antaki, MDCM, FRCSC, DABO. Cole Eye Institute, Cleveland Clinic, Ohio, USA. Email: antakif@ccf.org

**ORCiD of authors:** Fares Antaki (0000-0001-6679-7276), David Mikhail (0009-0009-0831-1915), Daniel Milad (0000-0002-0693-3421), Danny A. Mammo (0000-0002-7496-5118), Sumit Sharma (0000-0001-5769-0717), Sunil K. Srivastava (0000-0002-0398-8806), Bing Yu Chen (0000-0003-4049-7528), Samir Touma (0000-0002-6365-0946), Mertcan Sevgi (0009-0003-8426-6534), Jonathan El-Khoury (0000-0003-3186-2351), Pearse A Keane (0000-0002-9239-745X), Qingyu Chen (0000-0002-6036-1516, Yih Chung Tham (0000-0002-6752-797X), Renaud Duval (0000-0002-3845-3318)

**Financial Disclosures:** Dr. Antaki is an equity owner in SIMA Surgical Intelligence Inc. Dr. Sharma serves as a consultant for 4DMT, Alimera, Abbvie, Apellis, Astellas, Bausch and Lomb, Clearside, Eyepoint, Harrow, Genetech/Roche, Kodiak, Merck, Regeneron, RegenXBio, Ripple, Volk, and Zeiss with contracted research support from Acelyrin, Gilead, Genetech/Roche, Santen, IONIS, Kodiak. Dr. Keane has acted as a consultant for insitro, Retina Consultants of America, Roche, Boehringer-Ingleheim, and Bitfount and is an equity owner in Cascader Ltd and Big Picture Medical. He has received speaker fees from Zeiss, Thea, Apellis, and Roche, and grant funding from Roche. He has received travel support from Bayer and Roche. He has attended advisory boards for Topcon, Bayer, Boehringer-Ingleheim, and Roche. Dr. Duval has acted as a consultant for Roche, Bayer and Apellis. The remaining authors have nothing to declare.

**Funding Support:** No funding was obtained for this work.

**Ethics Approval:** Ethics approval was not required for this work.

**Patient Consent:** Patient consent was not required as this work did not involve patients.

**Acknowledgements:** We thank the American Academy of Ophthalmology for generously granting us permission to use the underlying BCSC Self-Assessment Program materials. Dr. Antaki is supported by an AI in Healthcare Fellowship Bursary by the CHUM Foundation and the Quebec Government. Dr. B. Chen is supported by an NIH StrokeNet training grant and Cleveland Clinic Caregiver Catalyst Grant. Dr. Q. Chen is supported by NIH grants R01LM014604 and R00LM014024. Dr. Keane is supported by a UK Research & Innovation Future Leaders Fellowship (MR/T019050/1), Moorfields Eye Charity with The Rubin Foundation Charitable Trust (GR001753), and an Alcon Research Institute Senior Investigator Award.

**Word Count:** 3421

# Key Points

**Question:** What are the capabilities of OpenAI's GPT-5 series compared with prior reasoning large language models on ophthalmology-specific multiple-choice questions?

**Findings:** In this cross-sectional study of three GPT-5 models (four reasoning efforts) and 3 prior reasoning LLMs on 260 AAO Basic Clinical Science Course questions, GPT-5-high achieved the highest accuracy (96.5%), outperforming all others. GPT-5-high ranked first in rationale quality, and cost-accuracy analysis identified several Pareto-efficient configurations, with GPT-5-mini-low as the best low-cost option.

**Meaning:** GPT-5 with high reasoning effort delivers near-perfect accuracy and top rationale quality, while lower-cost GPT-5 configurations maintain strong performance for high-accuracy and budget-conscious applications.

# Abstract

**Importance:** Novel large language models (LLMs) such as GPT-5 integrate advanced reasoning capabilities that may enhance performance on complex medical question-answering tasks. For this latest generation of reasoning models, the configurations that maximize both accuracy and cost-efficiency have yet to be established.

**Objective:** To evaluate the performance and cost-accuracy trade-offs of OpenAI's GPT-5 compared to previous generation LLMs on ophthalmological question answering.

**Design, Setting, and Participants:** In August 2025, 12 configurations of OpenAI's GPT-5 series (three model tiers across four reasoning effort settings) were evaluated alongside o1-high, o3-high, and GPT-4o, using 260 closed-access multiple-choice questions from the AAO Basic Clinical Science Course (BCSC) dataset. The study did not include human participants.

**Main Outcomes and Measures:** The primary outcome was accuracy on the 260-item ophthalmology multiple-choice question set for each model configuration. Secondary outcomes included head-to-head ranking of configurations using a Bradley-Terry (BT) model applied to paired win/loss comparisons of answer accuracy, and evaluation of generated natural language rationales using a reference-anchored, pairwise LLM-as-a-judge framework. Additional analyses assessed the accuracy-cost trade-off by calculating mean per-question cost from token usage and identifying Pareto-efficient configurations.

**Results:** The configuration GPT-5-high achieved the highest accuracy (0.965; 95% CI, 0.942–0.985), significantly outperforming all GPT-5-nano variants ($P < .001$), o1-high ($P = .04$), and GPT-4o ($P < .001$), but not o3-high (0.958; 95% CI, 0.931–0.981). The configuration GPT-5-high ranked first in accuracy (1.66x stronger than o3-high) and rationale quality (1.11x stronger than o3-high), as judged by a reference-anchored LLM-as-a-judge autograder. Cost-accuracy analysis identified multiple GPT-5 configurations on the Pareto frontier, with GPT-5-mini-low providing the most optimal low-cost, high-performance configuration.

**Conclusions and Relevance:** This study benchmarks the GPT-5 series on a high-quality ophthalmology question-answering dataset, demonstrating that GPT-5 with high reasoning effort achieved near-perfect accuracy and outperformed prior reasoning LLMs. This study also introduces an autograder framework for scalable, automated evaluation of LLM-generated answers against reference standards in ophthalmology.

# Introduction

Large language models (LLMs) have rapidly emerged as powerful tools in medicine, capable of answering complex questions and assisting with clinical decision support.[1] In ophthalmology, these models have been extensively benchmarked on specialty-specific knowledge tasks, reflecting growing interest in their potential utility.[2] Many studies have evaluated LLM performance on ophthalmology board-style examinations and clinical case scenarios, establishing baseline accuracies and identifying strengths and limitations in their medical reasoning.[3–8]

Numerous generations of models have emerged since the first evaluations of LLMs in ophthalmology.[9] More recently, *reasoning models* (systems that employ step-by-step logical processing before generating an output) have demonstrated improved performance in ophthalmological question answering. In a recent study, Srinivasan et al found that OpenAI's o1 model outperformed earlier-generation models such as GPT-4o in accuracy and usefulness.[10] Strong performance has also been reported with other reasoning models, including Gemini 2.0 Flash-Thinking (Google) and the open-source DeepSeek-R1.[4,11]

In August 2025, OpenAI released GPT-5, a new series of reasoning models intended to represent the state-of-the-art and establish a new performance frontier.[12] Accordingly, each new generation of models warrants reassessment to determine whether improvements are sustained and to identify any associated trade-offs, including potential new failure modes. Ideal evaluations should use rigorously curated datasets that minimize the risk of training data contamination.[13] Prior evaluations using a closed-access sample of the American Academy of Ophthalmology's Basic and Clinical Science Course (BCSC) dataset benchmarked GPT-3.5 at 55.8% accuracy and GPT-4 at 75.8%.[3]

In this study, we assessed the performance of the latest GPT-5 models on the BCSC dataset. We tested all three GPT-5 model variants (spanning different size/cost tiers) across four reasoning effort settings. The primary outcome was the overall response accuracy of each configuration. Secondary outcomes included a head-to-head comparison of model configurations, a qualitative comparison of the LLM responses with the BCSC ground truth explanations, and a cost analysis of model usage at each configuration.

# Methods

## The BCSC dataset

In August 2025, we obtained written permission from the AAO to use a set of 260 questions, sampled in 2023 from the 4,458-question pool in the BCSC Self-Assessment Program, which we had used in prior studies.[3,7] Because the dataset is behind a paywall and not publicly accessible, it serves as a consistent, closed-access benchmark for evaluating LLM performance. The dataset contains 20 text-only questions from each of the 13 ophthalmology subspecialties, as defined by the BCSC curriculum. The characteristics of the BCSC question set, including distribution by examination section, cognitive level, and difficulty, have been described previously.[7] Briefly, questions were labeled as low cognitive level if they primarily assessed factual recall and as high cognitive level if they required data interpretation or patient management. Difficulty was defined using the percentage of correct responses by human test-takers.

## GPT-5 series of models

GPT-5 is a series of LLMs released by OpenAI (San Francisco, California) on August 7, 2025.[12] Compared with GPT-4 and other OpenAI models, GPT-5 demonstrated superior performance on academic and human-evaluated benchmarks, including those in health.[14] In evaluations using HealthBench,[15] the GPT-5 models outperformed models like OpenAI GPT-4o, o1, o3, and o4-mini with higher accuracy and fewer hallucinations.[14] Three GPT-5 models were available for testing in the application programming interface (API): GPT-5 (best model), GPT-5-mini (faster and more cost-efficient), and GPT-5-nano (fastest and most cost-efficient).[16] OpenAI released GPT-5 as a unified system in which queries are processed by the specified main model, with more complex tasks automatically routed to a higher-reasoning capacity "thinking" version of that model. The choice between the main and thinking versions is made by a real-time routing mechanism based on query complexity. In the API, we were provided direct access to the thinking model. To establish a baseline, we tested the OpenAI o1 and o3 models with high reasoning effort. We also tested GPT-4o with a temperature setting of 0.3.

## Adjusting GPT-5's reasoning effort

The OpenAI Responses API allows adjustment of "reasoning effort", a parameter that controls the generation of reasoning tokens, which are internal tokens used by the model for processing and planning, before producing the final output. Higher effort settings generally result in more reasoning tokens and longer inference times but may improve accuracy and performance on challenging tasks, such as scientific or health benchmarks. The effort can be set to: minimal, low, medium, or high. The optimal GPT-5 reasoning effort for medical question answering has not yet been defined. We therefore tested GPT-5 on all four reasoning effort settings. For ease of reference, we refer to the three GPT-5 models with different reasoning efforts as "configurations" and label them, for example, as GPT-5-low, GPT-5-mini-medium, or GPT-5-nano-high.

## Prompting strategy

The questions were preserved in their original multiple-choice format, consisting of one correct option and three distractors. A zero-shot prompting strategy was used consistent with our prior work, as this approach most closely mirrors how humans answer test questions.[3] For each question, the model received: (i) a system instruction enforcing a structured output and brevity of justification, and (ii) the question text as the user message. The system instruction was:

> Answer this question. Return strict JSON with two keys: 'answer' and 'rationale'. 'answer' MUST be a SINGLE CAPITAL LETTER (A–D) only. 'rationale' MUST be a ONE-SENTENCE concise justification for your answer (no step-by-step).

We varied the *reasoning_effort* parameter across the four levels and otherwise used the API defaults. The letter <answer> was used to determine accuracy by comparing to the ground truth provided by the BCSC. The <rationale> natural language justification was used for secondary analyses. Token usage (input, output and reasoning) were logged for cost analyses.

## Multiple-choice accuracy evaluations

The primary outcome was to determine the accuracy of the different model configurations on the BCSC dataset. For each model configuration, results were based on a single pass, given our earlier findings with GPT-3.5 demonstrating substantial to almost perfect repeatability.[3] Before conducting the primary analyses, we evaluated all 12 GPT-5 configurations (3 models × 4 reasoning efforts) on 10% of the dataset (two questions per section, total of 26 questions) as a preliminary screening step to exclude the lowest-performing configurations. All three GPT-5 models with minimal reasoning effort performed the worst and were excluded from further analyses (**Supplemental Table 1**).

## Multiple-choice ranking evaluations

Secondary outcomes included a head-to-head comparison of model configurations. We implemented a head-to-head LLM 'arena' to perform paired comparisons of configurations for each question and derive a global ranking. For each question, every model configuration produced a letter response that was marked correct or incorrect against the answer key. We converted these outcomes into paired comparisons between configurations: a "win" was recorded when one configuration was correct and the other was incorrect. For each pair of model configurations, we report a head-to-head win rate.

We then fit a Bradley-Terry (BT) model, which assigns each configuration a positive skill parameter reflecting how often it tends to win against others on discordant items.[17] We report a BT skill metric which can be interpreted as the configuration's share of the total "ability" in the pool. The relative differences between skill values indicate how much stronger one configuration is compared with another in head-to-head performance. For example, if one configuration has a skill of 0.30 and another has 0.15, the first is estimated to be about twice as likely to produce a correct answer in direct comparisons across the question set.

## Autograder evaluation using LLM-as-a-judge

Secondary outcomes also included an evaluation of natural language justifications (<rationale>) generated by the different model configurations. We employed a reference-anchored, pairwise judging framework using an autograder (LLM-as-a-judge). This approach is scalable, greatly reduces the labor required for human grading and helps mitigate potential human bias.[18]

The judge model was provided with the question for context, the reference ground truth written by BCSC experts, and pairs of masked rationales generated by the different model configurations. To build the autograder, we used o4-mini with a custom prompt adapted from OpenAI's "Model Scorer" from the "Evaluations" dashboard. The autograder was instructed to identify salient facts from the BCSC reference, compare each pair of rationales, and select a winner or declare a tie. The detailed prompt is shown below:

> You judge two single-sentence rationales generated while answering an ophthalmology multiple-choice question against an authoritative reference. Use the QUESTION only for context; grade alignment to the REFERENCE_EXPLANATION.
>
> QUESTION: {{question_text}}
>
> REFERENCE_EXPLANATION: {{bcsc_reference_text}}
>
> RATIONALE_A: {{rationale_a}}
>
> RATIONALE_B: {{rationale_b}}
>
> Silently extract 2–4 salient facts from the reference.
>
> Choose the rationale that better matches those facts.
>
> Major contradictions or wrong mechanism/pathology are fatal.
>
> Return ONLY this one-line JSON:
>
> {"winner":"A"|"B"|"tie"}

After identifying the configurations with the highest multiple-choice accuracy, we selected the best-performing GPT-5 configuration (GPT-5-high) along with o1-high, o3-high, and GPT-4o to represent different model generations for this secondary analysis. For every question, 6 pairwise comparisons were performed, for a total of 1560. We randomized A/B assignment per pair to mitigate position bias.[19] Similar to the accuracy-based ranking, we report a head-to-head win rate and then fit a BT model to estimate each configuration's "skill", then rank configurations from highest to lowest based on these values.

## Accuracy-cost analysis

To determine which configurations are most beneficial, we plotted configuration-level accuracy against mean cost per question. We measured the mean token usage per configuration (input and output tokens) and converted these into a mean cost per question using OpenAI's per-million-token pricing at the time of analysis.[20] We plotted accuracy against cost for each configuration and identified the Pareto-efficient configurations. A configuration was considered Pareto-efficient if there was no other configuration with both higher accuracy and lower cost. The set of such non-dominated points forms the Pareto frontier.

## Statistical analysis

For the primary outcome, accuracy was compared across configurations using Cochran's Q test. Post-hoc pairwise comparisons between the best-performing configuration and all others were conducted with McNemar's exact test, applying Holm correction for multiple comparisons. For secondary analyses, we calculated win rates and visualized results in heatmaps, where values greater than 0.50 indicate the row model won more often and values less than 0.50 indicate it lost more often. BT models were fitted to estimate relative skill, and configurations were ranked by descending BT skill values. Confidence intervals for accuracy and BT skill metrics were estimated via bootstrapping with 1,000 iterations. Accuracy-cost scatterplots were generated with mean cost per question plotted on a base-10 logarithmic scale to accommodate the large dynamic range in costs across configurations. All analyses were conducted in Python, version 3.13.5. Statistical significance was set at $P < .05$.

# Results

## GPT-5-high was the most accurate

GPT-5-high achieved the highest accuracy (0.965; 95% CI, 0.942–0.985), whereas GPT-5-nano-low had the lowest (0.773; 95% CI, 0.723–0.823). Older generation reasoning models achieved high scores: o1 scored 0.927 (95% CI, 0.888–0.958) and o3 achieved 0.958 (95% CI, 0.931–0.981). Accuracy differed significantly across configurations (Q = 228.56; P < .001). In pairwise comparisons, GPT-5-high outperformed all GPT-5-nano configurations (P < .001), as well as o1 (P = .04), and GPT-4o (P < .001), but not o3 (P = 0.87). The results are summarized in **Table 1**. We provide the accuracy per cognitive level and question difficulty in **Supplementary Tables 2-3.**

**Table 1. Accuracy of different model configurations at varying reasoning efforts**

| Model | Reasoning effort | | |
|---|---|---|---|
| | Low | Medium | High |
| GPT-5 | 0.950 [0.923–0.973] | 0.954 [0.927–0.977] | **0.965 [0.942–0.985]** |
| GPT-5-mini | 0.927 [0.896–0.958] | 0.942 [0.911–0.969] | 0.942 [0.912–0.969] |
| GPT-5-nano | 0.773 [0.723–0.823] | 0.823 [0.781–0.865] | 0.831 [0.785–0.873] |
| o1 | | | 0.927 [0.888–0.958] |
| o3 | | | 0.958 [0.931–0.981] |
| GPT-4o | 0.865 [0.823–0.904] | | |

## GPT-5-high ranked first in head-to-head accuracy comparisons

**Figure 1** shows head-to-head accuracy win rates between model configurations. Overall rankings for accuracy skill as determined by BT modeling are shown in **Figure 2**. The top configuration was GPT-5-high (0.270; 95% CI, 0.118–0.670), followed by o3-high (0.163; 95% CI, 0.055–0.420) and GPT-5-medium (0.127; 95% CI, 0.037–0.311). The GPT-5 nano models ranked last. Top-ranked GPT-5-high was estimated to be 1.66 times stronger than o3-high and 5.10 times stronger than o1-high in head-to-head comparisons. We provide accuracy win rate between models per cognitive level and question difficulty in **Supplemental Figures 1 and 2**.

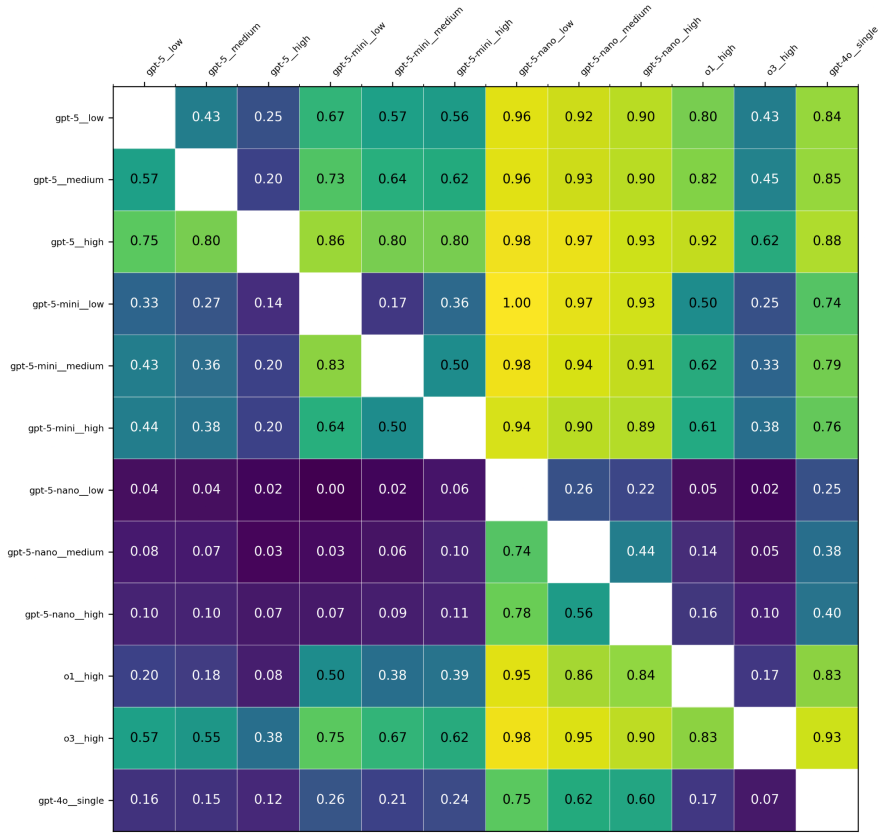

**Figure 1. Head-to-head accuracy win rates between model configurations.** Each cell shows the proportion of times the row model was correct and the column model was not. Values > 0.50 mean the row model won more often; values < 0.50 mean it lost more often.

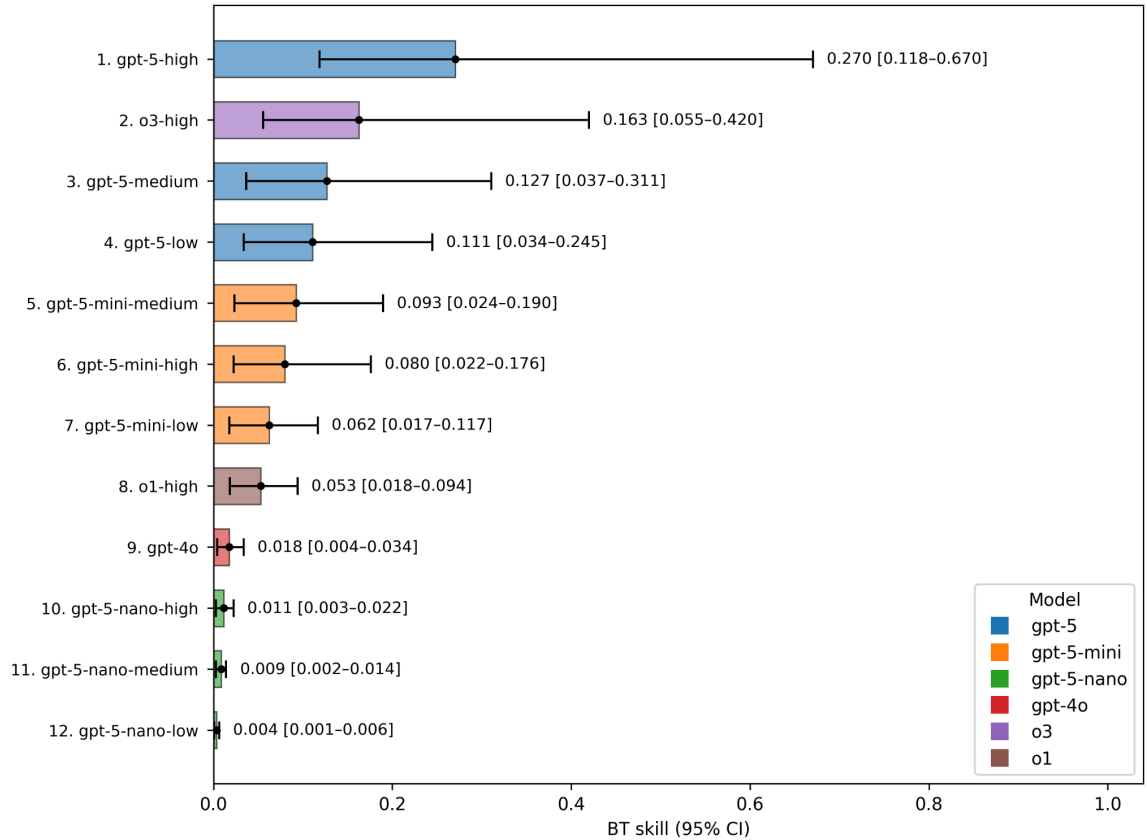

**Figure 2. Bradley-Terry estimated accuracy skills for all model configurations ranked from highest to lowest skill**

## GPT-5-high provided the best rationales

**Figure 3** shows head-to-head rationale win rates between the selected model configurations. Overall rankings for rationale skill as determined by BT modeling are shown in **Figure 4**. **Supplemental Table 4** shows specific examples of a GPT-5-high win over GPT-4o and an instance of a tie. The top configuration was GPT-5-high (0.358; 95% CI, 0.335–0.382), followed by o3-high (0.323; 95% CI, 0.303–0.345), o1-high (0.166; 95% CI, 0.152–0.180), and GPT-4o (0.153; 95% CI, 0.136–0.0.169). Top-ranked GPT-5-high was estimated to be marginally (1.11 times) stronger than o3-high, 2.16 times stronger than o1-high, and 2.34 times stronger than GPT-4o in head-to-head comparisons. We provide rationale win rates between models per cognitive level and question difficulty in **Supplemental Figures 3 and 4**.

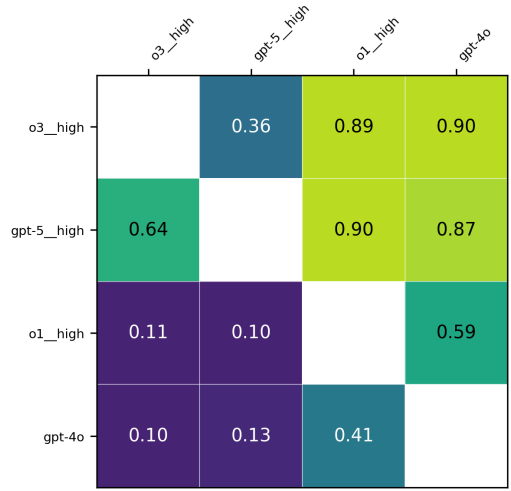

**Figure 3. Head-to-head rationale win rates between model configurations.** Each cell shows the proportion of times the row model was correct and the column model was not. Values > 0.50 mean the row model won more often; values < 0.50 mean it lost more often.

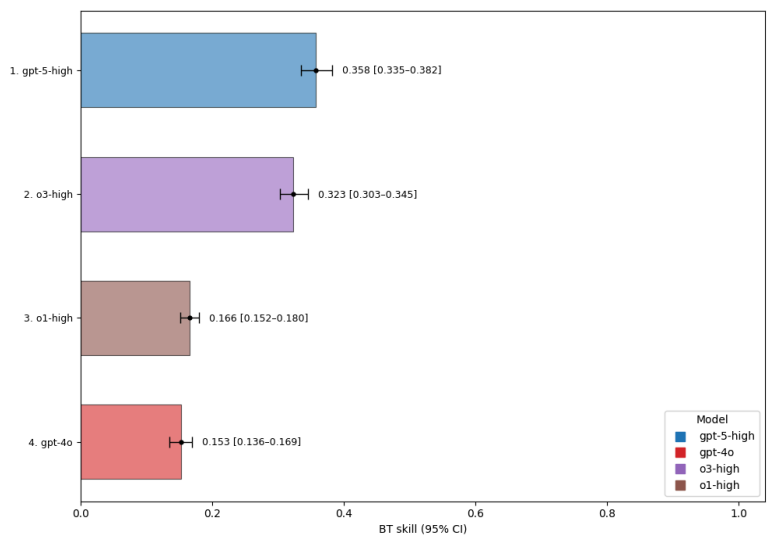

**Figure 4. Bradley-Terry estimated skills for rationale for all model configurations ranked from highest to lowest skill**

## GPT-5 series models gives the best cost-accuracy trade-offs

Across all configurations, the accuracy-cost Pareto frontier was formed by GPT-5-nano-low at the ultra-low-cost end and GPT-5-high at the high-cost end. GPT-5-mini-low was the pareto-optimal configuration with maximal accuracy and minimal cost. GPT-5-medium sits near o3-high on the frontier with a very similar cost-accuracy point. In contrast, o1-high costs more while delivering lower accuracy than cheaper frontier options. **Figure 5** shows the accuracy-cost points for all configurations. Detailed token usage and cost calculations are shown in **Supplemental Table 5**.

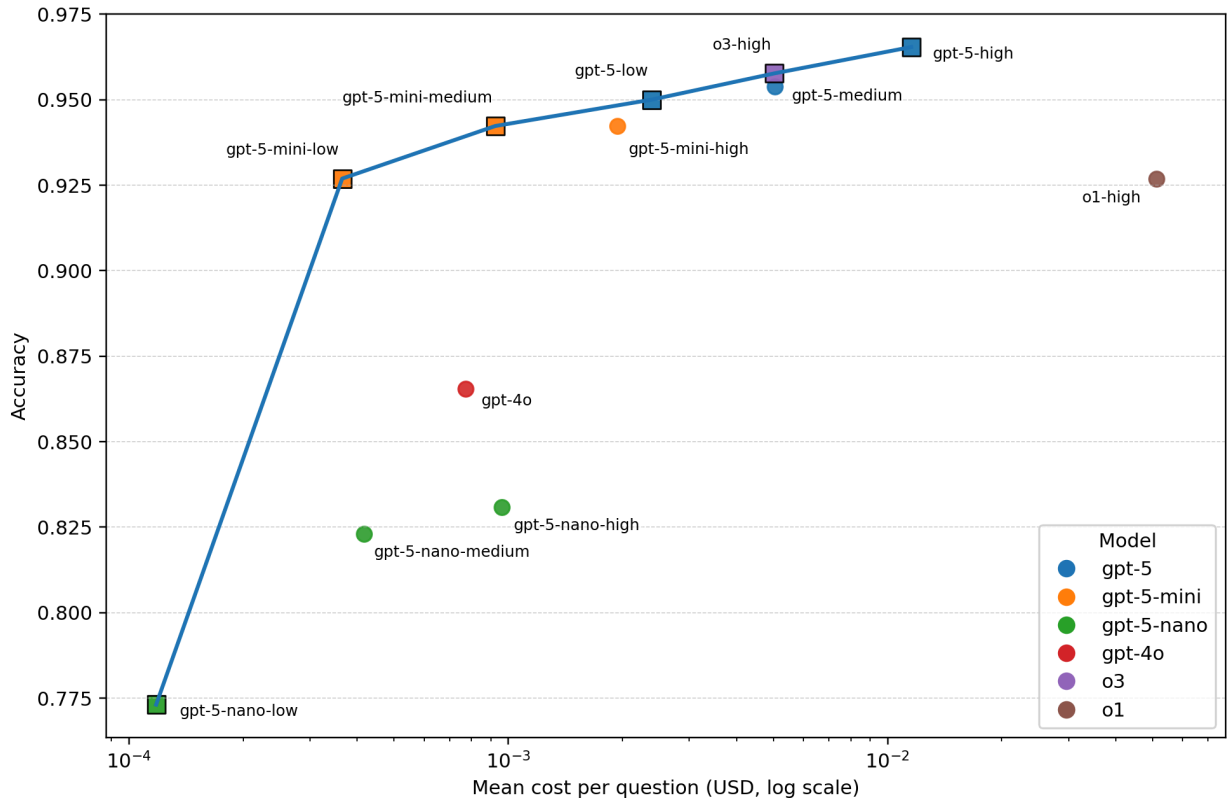

**Figure 5. Accuracy-cost trade-off across model configurations.** The x-axis is the mean cost per question (USD, log scale) and the y-axis is accuracy. Square marked configurations are Pareto-efficient, meaning no other configurations are both cheaper and more accurate. The line connects the Pareto frontier from lowest to highest cost.

## Discussion

LLMs are increasingly being considered for high-stakes applications in medicine, yet direct comparisons of accuracy, reasoning quality, and cost-efficiency across the latest model generations remain scarce. In this study, we evaluated the performance of novel GPT-5 series LLMs on ophthalmological medical question answering. We found that the GPT-5 model with high reasoning effort achieved near-perfect accuracy, ranked first in head-to-head comparisons with other models, and produced the highest-quality rationales. Cost-accuracy analysis identified GPT-5-mini with low reasoning effort as the Pareto-optimal configuration, delivering strong accuracy at minimal cost. These findings provide practical benchmarks to guide the implementation of frontier reasoning LLMs in real-world applications in ophthalmology.

Our preliminary analyses showed that GPT-5 configurations with minimal reasoning effort consistently underperformed across tiers and produced no reasoning tokens. The "minimal" setting may trigger automated routing to the non-thinking variant of the GPT-5 model, so these models were excluded from further analyses. In the full evaluation of the remaining nine GPT-5 configurations alongside three baselines, GPT-5 with high reasoning effort achieved the highest accuracy (0.965; 95% CI, 0.942–0.985), outperforming o1 and GPT-4o, but not o3. This near-perfect performance represents a substantial improvement from GPT-3.5 (55.8%) and GPT-4 (75.8%), both tested on the same dataset.[3] Our findings align with recent reports showing that GPT-5 outperforms GPT-4o in reasoning and understanding scores on general medicine benchmarks.[21] This technical achievement may be attributed to several factors, including training on a broader and more recent corpus, enhanced data curation, improved reasoning through reinforcement learning, and more advanced alignment tuning.

To compare configurations beyond simple accuracy, we used the BT model, which estimates the relative "skill" of each configuration from head-to-head outcomes, similar to the Elo rating system in competitive chess. This provides a more nuanced measure than raw accuracy by quantifying how often one model outperforms another.[22–25] In addition to testing answer accuracy, we ranked models on the quality of their generated rationales or justifications. To do so, we conducted a reference-anchored, pairwise evaluation in which an LLM-as-a-judge selected the better answer. This approach is far less labor-intensive than large-scale human grading, offers objectivity, scalability, and consistency.[26] Such autograding approaches have not, to our knowledge, been routinely applied in ophthalmology research, which has typically relied on metric-based evaluations (e.g., ROUGE and METEOR) that assess text similarity but may miss conceptual alignment and clinical nuance.[10,11]

In head-to-head comparisons, GPT-5 with high reasoning effort ranked first for both answer accuracy and rationale quality. For accuracy, it was estimated to be 1.66 times stronger than o3-high and 5.10 times stronger than o1-high. For rationale quality, it was only 1.11 times stronger than o3-high but was 2.16 times stronger than o1-high, and 2.34 times stronger than GPT-4o. These findings are impressive because o1 had already shown strong performance in ophthalmology.[4,11] Our previous work found that o1 achieved the highest accuracy (0.877) on large, open-access ophthalmology datasets such as MedMCQA, outperforming non-reasoning OpenAI models as well as Llama 3-8B (Meta) and Gemini 1.5 Pro (Google).[10] It also achieved the highest accuracy (0.882) on curated combinations of datasets such as the BEnchmarking LLMs for Ophthalmology (BELO) compared to DeepSeek-R1 and other models.[13]

GPT-5 with high reasoning effort consistently outperformed all other models in both accuracy and rationale quality across question cognitive levels and difficulties. We present an example of a high-cognitive-level question from the cornea section on the management of uveal prolapse during open-globe repair (**Supplemental Table 4**). The ground truth specifies that uveal tissue should be reposited, and only necrotic or contaminated tissue should be excised. GPT-5 with high reasoning effort accurately differentiated between when to reposition and when to resect the tissue. In contrast, GPT-4o prematurely recommended resection. The autograder correctly awarded the win to GPT-5. In general, complex clinical cases may benefit from models with advanced reasoning capabilities which are able to generate nuanced, context-aware responses.

The cost-accuracy analysis demonstrated that GPT-5 series models offered the most favorable trade-offs, with the Pareto frontier spanning from GPT-5-nano-low at the ultra-low-cost end to GPT-5-high at the high-accuracy end. The configuration o1-high was both more expensive and less accurate than several cheaper frontier options. GPT-5-mini-low was the Pareto-optimal configuration, combining maximal accuracy with minimal cost. This configuration may be advantageous because it is faster than the main GPT-5 model, making it suitable for applications that require rapid responses, such as patient triage or real-time chatbot interactions.

This study has limitations. First, the multiple-choice format does not fully replicate real-world clinical decision-making; however, our objective was to identify optimal configurations using a closed dataset to facilitate comparison with prior work. Repeated testing of LLMs with the same questions may risk leakage into training data, either through our use or by other researchers employing the BCSC dataset. Since initiating this line of work, we have taken measures to mitigate this risk: in our ChatGPT study, conducted before API access was available, we requested data deletion from OpenAI and obtained written confirmation that these data would not be used for model training;[7] in our GPT-4 study, we used the API, which, by design, does not contribute user inputs to training.[3] Future work will evaluate the best-performing configurations identified in this study on more complex, free-response ophthalmology cases and compare their performance with that reported in prior studies.[4–6]

Second, to assess rationale quality, we employed, to our knowledge for the first time in ophthalmology LLM research, an LLM-as-a-judge approach. This method carries potential biases, as performance depends on the choice of grading model, the prompt design, and the evaluation format. We mitigated verbosity bias by using single-sentence justifications and the position bias by using randomized A/B assignment per pair. Future work could explore longer-form rationales within the same autograding framework. Finally, the cost analysis was based on token usage and pricing at the time of testing; changes in API pricing, model efficiency, routing behavior, or dataset characteristics could alter the relative cost-accuracy trade-offs observed here. Performance and cost may also differ on open-ended, image-based, or multi-step clinical scenarios not represented in our dataset.

In conclusion, among the configurations tested, GPT-5 with high reasoning effort achieved the highest accuracy and rationale quality, outperforming most prior-generation reasoning models while offering favorable cost-accuracy trade-offs. Pareto analysis demonstrated that users can achieve high accuracy at a low cost with the GPT-5-mini models. These findings provide a practical framework for selecting LLM configurations in ophthalmology, whether for educational tools, research applications, or early-stage clinical decision-support systems. Future work should validate these results on complex, free-response, and multimodal ophthalmology tasks, explore the generalizability of LLM-as-a-judge for rationale evaluation, and assess real-world cost and deployment considerations.

# Supplemental Materials

**Supplemental Table 1. Comparison of GPT-5 model configurations at different reasoning efforts on the preliminary dataset (n=20).** GPT-5 models with minimal reasoning effort had the lowest accuracy and were excluded from the main analyses.

| Model | Reasoning effort | | | |
| --- | --- | --- | --- | --- |
| | Minimal | Low | Medium | High |
| GPT-5 | 0.846 [0.692–0.962] | 1.000 [1.000–1] | 0.962 [0.885–1] | 0.923 [0.808–1] |
| GPT-5-mini | 0.885 [0.768–1] | 0.923 [0.808–1] | 0.962 [0.885–1] | 0.923 [0.808–1] |
| GPT-5-nano | 0.692 [0.500–0.846] | 0.923 [0.808–1] | 0.885 [0.769–1] | 0.885 [0.769–1] |

**Supplemental Table 2. Accuracy of model configurations at different cognitive levels**

| Cognitive Level | Model | Reasoning Effort | Accuracy [95% CI] |
|---|---|---|---|
| **High** | GPT-5 | Low | 0.915 [0.858–0.962] |
| | | Medium | 0.925 [0.877–0.972] |
| | | High | 0.943 [0.896–0.981] |
| | GPT-5-mini | Low | 0.877 [0.811–0.934] |
| | | Medium | 0.896 [0.840–0.943] |
| | | High | 0.906 [0.849–0.953] |
| | GPT-5-nano | Low | 0.708 [0.613–0.792] |
| | | Medium | 0.783 [0.708–0.858] |
| | | High | 0.783 [0.708–0.858] |
| | o1 | High | 0.887 [0.821–0.943] |
| | o3 | High | 0.943 [0.896–0.981] |
| | GPT-4o | N/A | 0.811 [0.726–0.878] |
| **Low** | GPT-5 | Low | 0.974 [0.948–0.994] |
| | | Medium | 0.974 [0.948–0.994] |
| | | High | 0.981 [0.955–1.000] |
| | GPT-5-mini | Low | 0.961 [0.929–0.987] |
| | | Medium | 0.974 [0.948–0.994] |
| | | High | 0.968 [0.942–0.994] |
| | GPT-5-nano | Low | 0.818 [0.760–0.883] |
| | | Medium | 0.851 [0.792–0.909] |
| | | High | 0.864 [0.812–0.916] |
| | o1 | High | 0.955 [0.922–0.987] |
| | o3 | High | 0.968 [0.935–0.994] |
| | GPT-4o | N/A | 0.903 [0.851–0.942] |

**Supplemental Table 3. Accuracy of model configurations at different difficulty categories**

| Difficulty Category | Model | Reasoning Effort | Accuracy [95% CI] |
|---|---|---|---|
| Easy | GPT-5 | Low | 0.667 [0.333–0.889] |
| | | Medium | 0.667 [0.333–0.889] |
| | | High | 0.778 [0.444–1.000] |
| | GPT-5-mini | Low | 0.556 [0.222–0.889] |
| | | Medium | 0.556 [0.222–0.889] |
| | | High | 0.667 [0.333–0.889] |
| | GPT-5-nano | Low | 0.444 [0.111–0.778] |
| | | Medium | 0.556 [0.222–0.889] |
| | | High | 0.556 [0.222–0.889] |
| | o1 | High | 0.556 [0.222–0.889] |
| | o3 | High | 0.667 [0.333–0.889] |
| | GPT-4o | N/A | 0.556 [0.222–0.889] |
| Moderate | GPT-5 | Low | 1.000 [1.000–1.000] |
| | | Medium | 0.986 [0.966–1.000] |
| | | High | 0.993 [0.979–1.000] |
| | GPT-5-mini | Low | 0.973 [0.945–0.993] |
| | | Medium | 0.993 [0.979–1.000] |
| | | High | 0.979 [0.952–1.000] |
| | GPT-5-nano | Low | 0.849 [0.795–0.911] |
| | | Medium | 0.877 [0.815–0.925] |
| | | High | 0.904 [0.856–0.945] |
| | o1 | High | 0.979 [0.952–1.000] |
| | o3 | High | 0.986 [0.966–1.000] |
| | GPT-4o | N/A | 0.952 [0.911–0.986] |
| Difficult | GPT-5 | Low | 0.905 [0.848–0.962] |
| | | Medium | 0.933 [0.886–0.981] |
| | | High | 0.943 [0.895–0.981] |
| | GPT-5-mini | Low | 0.895 [0.829–0.952] |
| | | Medium | 0.905 [0.848–0.952] |
| | | High | 0.914 [0.857–0.962] |
| | GPT-5-nano | Low | 0.695 [0.600–0.781] |
| | | Medium | 0.771 [0.686–0.848] |
| | | High | 0.752 [0.657–0.829] |
| | o1 | High | 0.886 [0.819–0.943] |
| | o3 | High | 0.943 [0.895–0.981] |
| | GPT-4o | N/A | 0.771 [0.686–0.848] |

**Supplemental Table 4. Examples of a GPT-5-high win and tie versus GPT-4o**

| Question | BCSC Ground truth and model responses |
|---|---|
| What is the best way to manage uveal prolapse when it occurs during an acute open-globe repair?<br>A. reposit tissue<br>B. resect tissue<br>C. repair/reconstruct tissue<br>D. cauterize tissue | **BCSC ground truth**<br>If uvea or retina (seen as translucent, tan tissue with extremely fine vessels) protrudes, the surgeon can reposit it, using a gentle sweeping technique through a separate limbal incision, with the assistance of viscoelastic injection to keep the anterior chamber formed. Only in cases of frankly necrotic, macerated, or contaminated tissue should prolapsed uveal tissue be excised. Cauterization of uveal tissue would not be appropriate. Reconstruction of iris tissue, using techniques such as the McCannel suture, may be indicated in a follow-up surgery, but not during the repair of an acute open globe.<br><br>**GPT-5-high rationale**<br>During primary open-globe repair, viable prolapsed uveal tissue should be gently reposited, with resection reserved only for devitalized or contaminated tissue.<br><br>**GPT-4o rationale**<br>Resecting the prolapsed uveal tissue is preferred to prevent further complications and infection<br><br>**Autograder vote =** GPT-5-high |
| A 6-month-old child presents with constant tearing, moderate mucopurulent discharge, and a white/quiet eye. What is the most likely diagnosis?<br>A. nasolacrimal duct (NLD) obstruction<br>B. infectious conjunctivitis<br>C. dacryoadenitis<br>D. canalicular dysgenesis | **BCSC ground truth**<br>Canalicular dysgenesis leads to impaired drainage of tears, resulting in epiphora without associated signs of dacryocystitis, such as redness, swelling, pain, or mucopurulence. Infectious conjunctivitis typically causes a red eye and presents with tearing and irritation. NLD obstruction, either as a primary congenital condition or secondary to edema of the nasal tissue during an upper respiratory illness, causes tearing. Presentation can range from minimal or no discharge to copious discharge and associated dacryocystitis. Dacryoadenitis is inflammation of the lacrimal gland.<br><br>**GPT-5 high rationale**<br>A white, quiet eye with constant tearing and mucopurulent discharge in an infant is classic for nasolacrimal duct obstruction rather than conjunctivitis or lacrimal gland pathology.<br><br>**GPT-4o rationale**<br>The symptoms of constant tearing and mucopurulent discharge with a white/quiet eye are characteristic of nasolacrimal duct obstruction in infants.<br><br>**Autograder vote =** tie |

**Supplemental Table 5. Token usage and cost calculations of model configurations at different reasoning efforts.** *Reasoning tokens are included in output tokens

| Model | Reasoning Effort | Accuracy | Mean Cost Per Question (USD) | Mean Input Tokens | Mean Output Tokens | Mean Reasoning Tokens* | Mean Total Tokens |
|---|---|---|---|---|---|---|---|
| GPT-5 | Low | 0.950 | 0.002 | 133.442 | 221.935 | 171.815 | 355.377 |
| GPT-5 | Medium | 0.954 | 0.005 | 133.442 | 486.988 | 437.415 | 620.431 |
| GPT-5 | High | 0.965 | 0.012 | 133.442 | 1136.800 | 1086.769 | 1270.242 |
| GPT-5-mini | Low | 0.927 | 0.000 | 133.442 | 165.896 | 115.692 | 299.338 |
| GPT-5-mini | Medium | 0.942 | 0.001 | 133.442 | 446.058 | 396.800 | 579.500 |
| GPT-5-mini | High | 0.942 | 0.002 | 133.442 | 952.123 | 902.646 | 1085.565 |
| GPT-5-nano | Low | 0.773 | 0.000 | 133.442 | 278.131 | 225.723 | 411.573 |
| GPT-5-nano | Medium | 0.823 | 0.000 | 133.442 | 1025.450 | 974.523 | 1158.892 |
| GPT-5-nano | High | 0.831 | 0.001 | 133.442 | 2389.754 | 2340.185 | 2523.196 |
| o1 | High | 0.927 | 0.051 | 133.442 | 820.338 | 785.969 | 953.781 |
| o3 | High | 0.958 | 0.005 | 133.442 | 595.388 | 547.446 | 728.831 |
| GPT-4o | N/A | 0.865 | 0.001 | 130.442 | 44.735 | 0.000 | 175.177 |

**Supplemental Figure 1. Head-to-head accuracy win rates between model configurations per cognitive level.** Each cell shows the proportion of times the row model was correct and the column model was not. Values > 0.50 mean the row model won more often; values < 0.50 mean it lost more often.

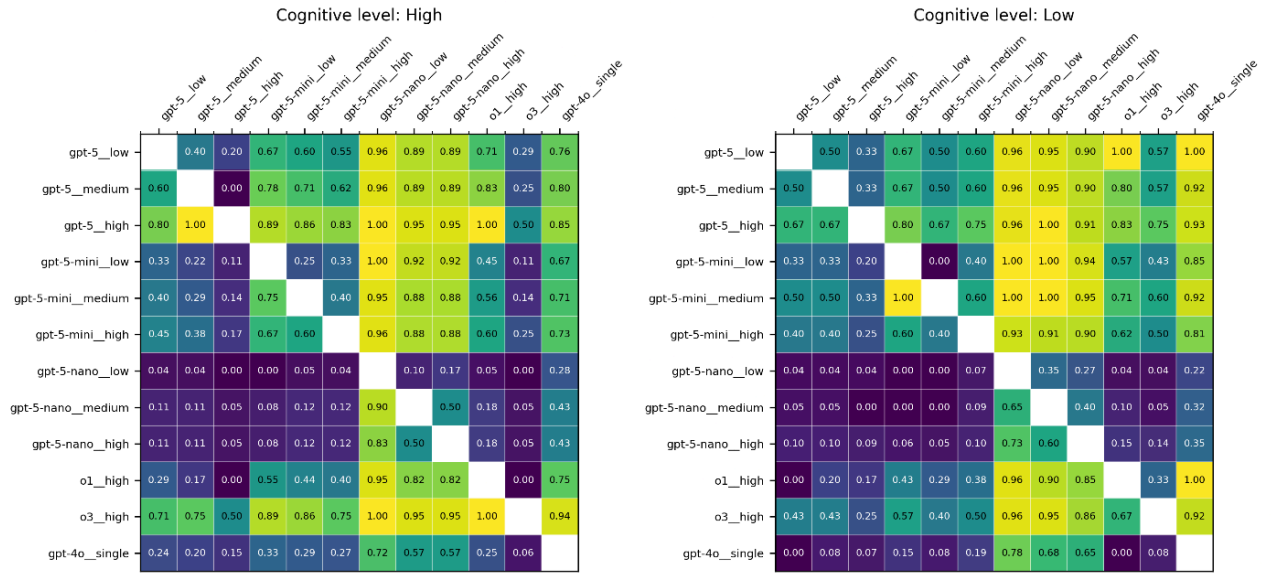

**Supplemental Figure 2. Head-to-head accuracy win rates between model configurations per question difficulty.** Each cell shows the proportion of times the row model was correct and the column model was not. Values > 0.50 mean the row model won more often; values < 0.50 mean it lost more often.

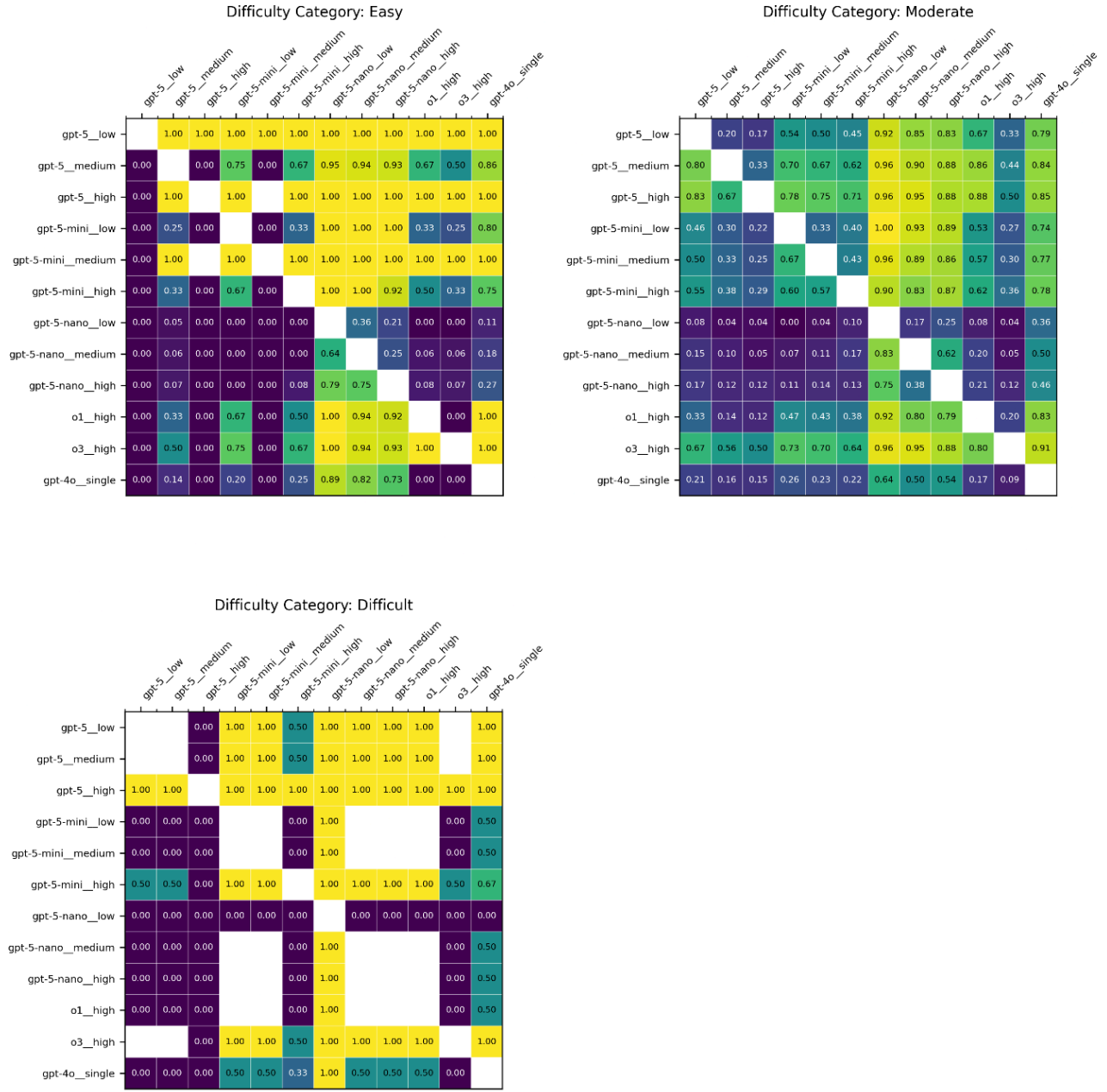

**Supplemental Figure 3. Head-to-head rationale win rates between model configurations per cognitive level.** Each cell shows the proportion of times the row model was correct and the column model was not. Values > 0.50 mean the row model won more often; values < 0.50 mean it lost more often.

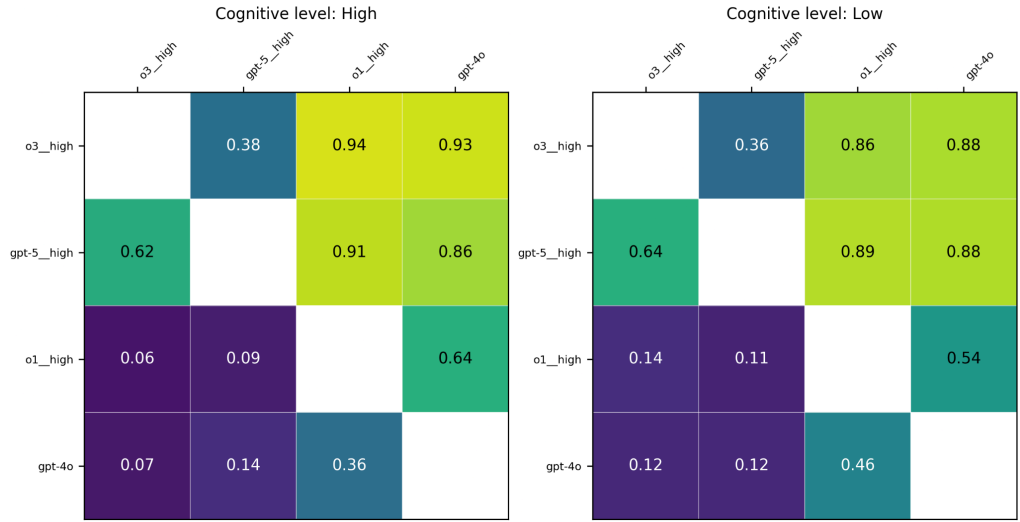

**Supplemental Figure 4. Head-to-head rationale win rates between model configurations per question difficulty.** Each cell shows the proportion of times the row model was correct and the column model was not. Values > 0.50 mean the row model won more often; values < 0.50 mean it lost more often.

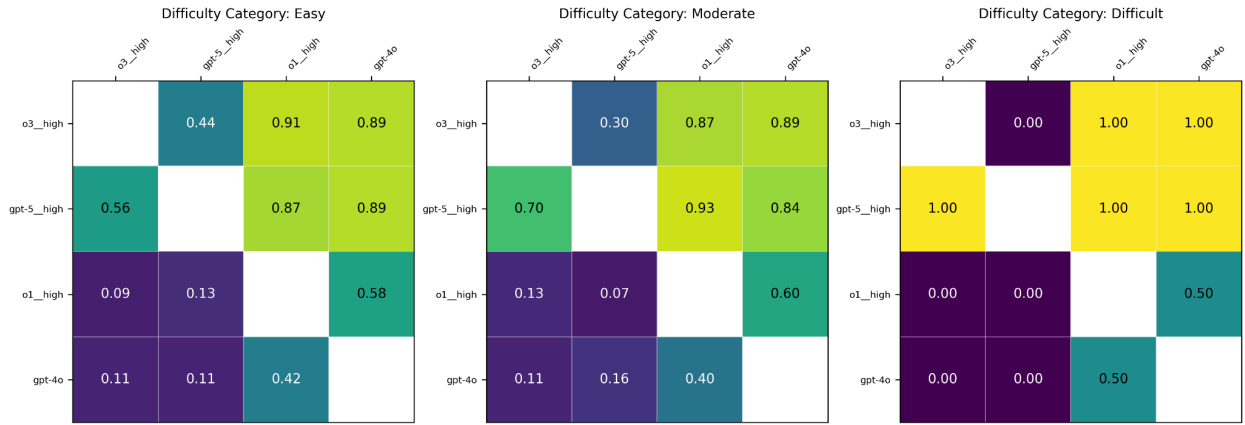